\pdfoutput=1

\documentclass[11pt]{article}

\usepackage[final]{acl}

\usepackage{times}
\usepackage{latexsym}

\usepackage[T1]{fontenc}

\usepackage[utf8]{inputenc}

\usepackage{microtype}

\usepackage{inconsolata}

\usepackage{graphicx}
\usepackage{tabularx}
\usepackage{booktabs}
\usepackage{adjustbox} 
\usepackage{amssymb}

\usepackage{float}
\usepackage{longtable}
\usepackage{amsmath}
\usepackage{subcaption}
\usepackage{multirow}
\usepackage{placeins}
\usepackage{comment}

\title{Reinforcement Learning for Neural Model Editing}

\author{
  \textbf{Shaivi Malik\textsuperscript{1}}
\\
\\
 \textsuperscript{1}Independent
}

\begin{document}
\maketitle
\maketitle
\begin{abstract}
Editing pretrained neural networks requires specialized algorithms tailored to specific objectives. Designing such algorithms is often time-consuming and demands significant effort. We present an exploratory framework that formulates neural model editing as a reinforcement learning problem, where agents modify models using only reward feedback. We introduce two environments: MaskWorld, where agents scale weights multiplicatively, and ShiftWorld, where agents apply additive weight updates. The reward function combines a utility-preservation objective with a task-specific editing objective, enabling agents to learn targeted modifications while maintaining overall model performance. We evaluate the framework on bias mitigation in text classification and machine unlearning in image classification, both of which traditionally rely on specialized algorithms. Our results show that the learned edits reduce forget set accuracy to 0\% while preserving over 90\% retain set accuracy in the unlearning setting. In the bias mitigation setting, the learned edits improve performance on an unbiased evaluation dataset by more than 7\% while maintaining general classification utility. These findings suggest that reinforcement learning can serve as a general framework for learning neural model editing policies from reward feedback. \footnote{Our code, data, and evaluation results are publicly available at \url{https://github.com/shaivimalik/hyperl}.}

\end{abstract} 

\section{Introduction}
Reinforcement learning (RL) has been studied for decades as a framework for sequential decision-making through interaction with an environment. In RL, an agent learns through trial and error by taking actions and receiving feedback in the form of rewards, enabling it to optimize its behavior over time \citep{sutton1998reinforcement}. Recent advances in RL have led to superhuman performance in games such as chess and Go, breakthroughs in protein structure prediction, and alignment of language models with human preferences \citep{jumper2021highly, silver2017mastering, ouyang2022training}.

Modern neural networks are first trained on large-scale datasets and then fine-tuned on downstream tasks of interest. However, modifying a pretrained model to achieve objectives, such as concept erasure, bias mitigation, or incorporating new knowledge, requires specialized algorithms \citep{hort2024bias, tian2025identifying, meng2022locating, gur2025precise}. Motivated by prior work on learned optimization and hypernetworks \citep{li2016learning, ha2016hypernetworks}, we ask whether neural model editing itself can be learned.
\begin{figure}[H]
  \centering
  \includegraphics[width=1\columnwidth]{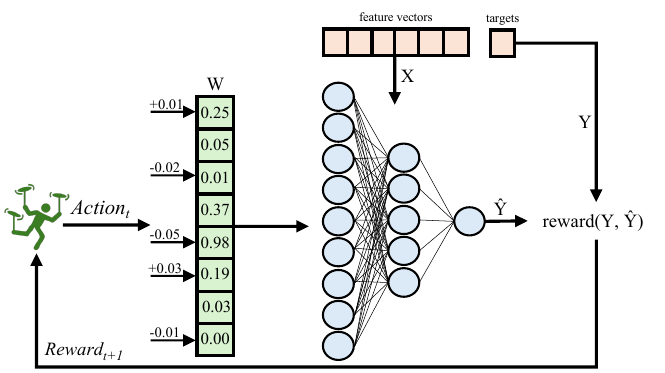}
  \caption{\textbf{Overview of our method.} In ShiftWorld, a reinforcement learning (RL) agent interacts directly with the weights of a neural network, modifying them based on rewards obtained on previous actions. The agent initially adds random numbers to the weights and gradually learns the weight modification that maximizes the reward. The reward function is designed to measure the extent to which the neural network exhibits the intended behavior.}
  \label{fig:one}
\end{figure}
In this study, we investigate the following research question: Can reinforcement learning agents learn to modify neural networks using only reward feedback? To explore this question, we create environments in which an RL agent interacts with a neural network by updating its weights and receiving rewards based on the performance of the neural network. We consider two kinds of weight updates: masking, where the agent scales a weight by a constant, and shifting, where the agent adds or subtracts a constant from a weight. At each timestep, the agent receives the current weights as observation and the reward resulting from the previous action; no gradients are computed or provided at any stage of the process. In particular, we investigate whether an agent can learn a policy for updating model weights to achieve goals that are difficult to construct datasets for and require specialized algorithms, such as unlearning, removal of spurious correlations, and improving fairness without degrading model utility.  

In Section~\ref{sec:method}, we formalize the environments and define the reward function as a combination of a utility term and a task-specific objective term. We evaluate our framework across multiple tasks and modalities: machine unlearning in convolutional neural networks for computer vision, and bias mitigation in text-based neural networks for natural language processing (Section~\ref{sec:exp_setup}). For machine unlearning, we evaluate the learned edits using accuracy on the retain and forget sets. For bias mitigation, we evaluate the learned edits on a dataset specifically designed to expose model biases (Section~\ref{sec:res}). Our framework achieved an accuracy of nearly 0\% on the forget set, demonstrating effective unlearning, while maintaining >90\% accuracy on the retain set. Our framework also improved performance on the unbiased dataset by more than 7\% in the bias mitigation task while maintaining general utility.
\section{Method}
\label{sec:method}

In our study, we create reinforcement learning environments for an agent to interact with the parameters of neural networks. The first environment, \textit{MaskWorld}, allows the agent to scale model weights by multiplying them with a constant. After each action, the agent receives the updated weights as the next observation and a reward reflecting the quality of the modification with respect to the target objective. The second environment, \textit{ShiftWorld}, instead allows additive updates, where the agent modifies weights by adding a constant. The reward function is task-dependent and is designed to capture the utility of the modified model and performance with respect to the editing objective. For example, in our machine unlearning setup, the reward is defined using a combination of accuracy on the retain set and accuracy on the forget set, encouraging the agent to preserve useful knowledge while selectively removing targeted information.

\subsection{Environments}
\noindent\textbf{MaskWorld.} In MaskWorld, an agent interacts with a neural network by modifying the weights of one layer of the neural network. The agent has the ability to scale the weights by a constant. The updated weights for layer \( l \) at timestep \( t \) are given by 
\( {W}^{(l)}_t = {W}^{(l)} \odot a_t \), where \( a_t \) is the action at timestep \( t \) and has the same dimensions as \({W}^{(l)}\). After performing an action, the agent receives a reward, as detailed in Section~\ref{subsec:reward}. \\
\noindent\textbf{ShiftWorld.} In ShiftWorld, similar to MaskWorld, an agent interacts with a neural network by modifying the weights of one layer of the neural network. The agent has the ability to shift the weights by a constant. The updated weights for layer \( l \) at timestep \( t \) are given by 
\( {W}^{(l)}_t = {W}^{(l)} + a_t \), 
where \( a_t \) is the action at timestep \( t \) and has the same dimensions as \({W}^{(l)}\). 

We allow the agent to modify the weights multiple times within a single episode. After each action, the agent receives a reward, defined in Section~\ref{subsec:reward}.  

For a weight matrix \(W \in \mathbb{R}^{m \times n}\), the policy is required to predict \(m \times n\) continuous actions at every step, which can be computationally prohibitive and lead to unstable policy optimization for high-dimensional layers. To resolve this, we model the agent’s actions using a formulation inspired by LoRA \citep{hu2022lora}. Instead of predicting an action of dimension \(m \times n\), the agent predicts two matrices \(A \in \mathbb{R}^{m \times r}\) and \(B \in \mathbb{R}^{r \times n}\). The updated weights at timestep \( t \) are then given by 
\( {W}^{(l)}_t = {W}^{(l)} + A \cdot B \) for ShiftWorld and \( {W}^{(l)}_t = {W}^{(l)} \odot A \cdot B \) for MaskWorld, where \( A \) and \( B \) are obtained from \( a_t \). This reduces the dimensionality of the action space, improving scalability. However, when the number of parameters is small, the policies can learn the optimal actions without this formulation. 

\subsection{Reward Function}
\label{subsec:reward}

We define the reward function as a weighted combination of two components: (i) an utility term that captures the general performance of the model, and (ii) an editing objective-specific term that quantifies the extent to which the modified model satisfies a desired property. The reward is defined as
\[
R = \alpha U(w) + \beta O(w)
\]
where \(U(w)\) denotes the utility of the model, \(O(w)\) denotes the quantity associated with the target objective, and \(\alpha\) and \(\beta\) coefficients control the trade-off between the two components.

For example, when the objective is to improve fairness, the utility term \(U(w)\) may correspond to classification accuracy, while the objective-specific term \(O(w)\) may be defined using a fairness metric such as equalized odds. In this setting, the reward function encourages modifications that improve fairness while preserving predictive performance.
\section{Experimental Setup}
\label{sec:exp_setup}
\subsection{Machine Unlearning}
\noindent\textbf{Problem Formulation.} In machine unlearning, given a neural network trained on dataset $D$ (the original model), with a retain set $D_r$ and a forget set $D_f$, the goal is to provide a set of weights for the neural network such that the resulting model (the unlearned model), unlearns the forget set $D_f$ while maintaining utility on the retain set $D_r$ and preserving broader generalization \citep{7163042, 9519428}.\\
\noindent\textbf{Task.} We train a convolutional neural network (CNN) on the MNIST dataset \citep{lecun1998gradient, lecun1998mnist}, which serves as the original model. The MNIST dataset \citep{lecun1998gradient, lecun1998mnist} consists of 70,000 grayscale images of handwritten digits (0–9). We task the agent with forgetting the class 7 while preserving performance on the remaining classes 1–6 and 8–9.\\
\noindent\textbf{Baselines.} We compare our framework against models modified using Random Masking, Random Shifting, and Gradient Ascent, as well as a model trained soley on the retain set.\\
\noindent\textbf{Reward Function.} We measure the utility of the model using accuracy on the retain set, i.e., classification performance on images corresponding to digits 0–6 and 8–9 and the editing objective as the accuracy on the forget set. We set \( \beta < 0\) to penalize performance on the forget set.
\[
\begin{aligned}
\text{Reward} =\;&
\alpha \left(
\frac{1}{|D_r|}
\sum_{i \in D_r}
\mathbf{1}(\hat{y}_i = y_i)
\right) \\
&+
\beta \left(
\frac{1}{|D_f|}
\sum_{j \in D_f}
\mathbf{1}(\hat{y}_j = y_j)
\right)
\end{aligned}
\]
\subsection{Bias Mitigation}
\noindent\textbf{Problem Formulation.} In bias mitigation, given a neural network trained on dataset $D$ (the original model), the goal is to provide a set of weights for the neural network such that the resulting model (the debiased model) reduces discriminatory or spurious behavior while maintaining predictive utility and preserving broader generalization. \\
\noindent\textbf{Task.} We train a neural network on the Jigsaw Toxic Comment Classification Challenge dataset \citep{jigsaw-toxic-comment-classification-challenge} to create a deliberately biased model. The Jigsaw dataset is designed for classification of toxic language in online comments. The Jigsaw dataset exhibits biases that induce spurious correlations in the trained model, causing it to associate words such as “black” with toxicity. We train the neural network to perform the binary classification task of predicting whether a given text sample is toxic or non-toxic. We use accuracy on the synthetic test set introduced by \citet{dixon2018measuring} as the measure of the bias exhibited by the model. We task the agent with reducing biases acquired during training while preserving predictive performance on the original classification task. \\
\noindent\textbf{Baselines.} We compare our framework against models modified using Random Masking, Random Shifting, and fine-tuning on an unbiased dataset.\\
\noindent\textbf{Reward Function.} We measure the utility of the model using accuracy on the Jigsaw Toxic Comment Classification Challenge dataset \citep{jigsaw-toxic-comment-classification-challenge} and measure how well the model achieves the bias mitigation objective using accuracy on the synthetic test set from \citet{dixon2018measuring}.
\subsection{Implementation Details}
The reinforcement learning policies were trained using Proximal Policy Optimization (PPO) \citep{schulman2017proximal} as implemented in Stable-Baselines3 \citep{stable-baselines3}. An MLP of [128, 128] neurons with ReLU activation was used to model the policy. Instead of showing the agent the entire dataset during a timestep, we sampled a different batch for each timestep. Each experiment was repeated with 15 different random seeds to ensure robustness of the results. We employ our LoRA-inspired mechanism for the machine unlearning task but deliberately omit it for the bias mitigation task, facilitating evaluation across both. Ablation studies analysing the effects of episode length, number of samples used to calculate the reward at each timestep, and reward function coefficients are provided in Appendix~\ref{subsec:ablation}. The complete implementation is available at \url{https://github.com/shaivimalik/hyperl}.

\begin{table*}[t]
\centering
\small
\setlength{\tabcolsep}{8pt}
\begin{tabular}{lcccc}
\toprule
\textbf{Model/Method}
& \textbf{Accuracy on Retain Set} $\uparrow$
& \textbf{Accuracy on Forget Set} $\downarrow$ \\
\midrule
Original Model &
0.9241 $\pm$ 0.0168	& 0.9073 $\pm$ 0.0433 \\

Random Masking &
0.8613 $\pm$ 0.0342	& 0.8217 $\pm$ 0.1764 \\

Random Shifting &
0.9215 $\pm$ 0.0133	& 0.9021 $\pm$ 0.0488 \\

\textbf{MaskWorld} &
\textbf{0.9375 $\pm$ 0.0066} & \textbf{0.0000 $\pm$ 0.0000} \\

\textbf{ShiftWorld} &
\textbf{0.9381 $\pm$ 0.0060} & \textbf{0.0001 $\pm$ 0.0003} \\

Gradient Ascent &
0.8641 $\pm$ 0.0375	& 0.0523 $\pm$ 0.0578 \\

Trained Solely on Retain Set &
0.9339 $\pm$ 0.0071	& 0.0000 $\pm$ 0.0000 \\
\bottomrule
\end{tabular}
\caption{Performance of the original model and the agent-modified model on the machine unlearning task. Higher accuracy on the retain set and lower accuracy on the forget set indicates better unlearning performance. The results show that our framework effectively identified and modified the appropriate weights to achieve the target objective while preserving performance on the retain set.}
\label{tab:unlearning}
\end{table*}

\begin{table*}[t]
\centering
\small
\setlength{\tabcolsep}{8pt}
\begin{tabular}{lcccc}
\toprule
\textbf{Model/Method}
& \textbf{Accuracy on General Dataset} $\uparrow$
& \textbf{Accuracy on Unbiased Dataset} $\uparrow$\\
\midrule
Original Model
& 0.7967 $\pm$ 0.0091 &	0.5447 $\pm$ 0.0177 \\

Random Masking
& 0.7954 $\pm$ 0.0121 &	0.5466 $\pm$ 0.0207 \\

Random Shifting
& 0.7903 $\pm$ 0.0138 &	0.5415 $\pm$ 0.0261\\

\textbf{MaskWorld}
& \textbf{0.7472 $\pm$ 0.0365} &	\textbf{0.6171 $\pm$ 0.0348} \\

\textbf{ShiftWorld}
& \textbf{0.7490 $\pm$ 0.0232} &	\textbf{0.6220 $\pm$ 0.0308} \\

Fine-tuning on Unbiased Dataset
& 0.7369 $\pm$ 0.0313 &	0.6061 $\pm$ 0.0253 \\
\bottomrule
\end{tabular}
\caption{Performance comparison of different methods
on the general and unbiased datasets. Higher accuracy on the unbiased dataset with minimal degradation on the general dataset indicates more effective bias mitigation. The learned edits successfully target weight modifications that satisfy the debiasing objective without drastically degrading accuracy on general data.}
\label{tab:bias}
\end{table*}

\section{Main Results}
\label{sec:res}
Table~\ref{tab:unlearning} presents the performance of the learned policies on the machine unlearning task. The editing objective of the modification is to eliminate information associated with class 7 while preserving overall model utility. The policies were trained using the hyperparameter values presented in Appendix~\ref{subsec:hyperparameter Values}, and performance is reported on the test set. We measured the accuracy of the original model and the modified model on the retain set (classes 1-6 and 8-9) and the forget set (class 7). The results demonstrate that policies learned through MaskWorld and ShiftWorld are effective at reducing information related to the forget set, reducing accuracy on class 7 to nearly 0\% in both cases. It is also noteworthy that despite the simplicity of the reward function, the learned policies preserved performance on the retain set, suggesting that the reward signal was sufficient to guide selective forgetting rather than weight degradation. 

More specifically, the original model achieved 92.41\% accuracy on the retain set. After modification, the retain set accuracy was 93.75\% for MaskWorld, and 93.81\% for ShiftWorld, representing an 1\% improvement over the original model. This suggests that the learned policy not only preserved knowledge about the retain set but also identified weight modifications that slightly improved performance. It is also interesting that the agents had access to only 15\% of the MNIST training set during policy optimisation, yet the modified models generalised well to the unseen test set. \footnote{Refer to Appendix~\ref{subsec:dataset} for details of the dataset splits.}

Table~\ref{tab:bias} reports the accuracy on test sets derived from the Jigsaw Toxic Comment Classification Challenge dataset \citep{jigsaw-toxic-comment-classification-challenge} (general dataset) and the synthetic test set from \cite{dixon2018measuring} (unbiased dataset). Accuracy on the Jigsaw test set reflects the model’s general utility, while accuracy on the unbiased dataset serves as a measure of bias removal; higher accuracy on the unbiased dataset indicates a reduction in model bias. We compare the performance of the original model, the model modified by the agent using the learned edits, and models obtained by fine-tuning, random masking and random shifting the final layer of the original model. The results show that the modified model improves accuracy on the unbiased dataset by 7\%, indicating that the policy successfully identified and adjusted weights associated with spurious correlations learned by the original model. Specifically, the accuracy on the unbiased dataset increased from 54.47\% to 61.71\% for MaskWorld and 62.20\% for ShiftWorld, while preserving classification performance on the general dataset at above 70\%. It is also important to note that these results were obtained by modifying the weights of the final layer. A promising direction for future work is extending this formulation to multi-agent settings, in which multiple agents collectively modify layers, potentially yielding stronger performance gains.

More broadly, our results suggest that reinforcement learning may provide a useful framework for learning model editing policies from reward feedback. We hope this work motivates future research on learned approaches to model editing, including applications to fairness, robustness, privacy, and interpretability. Importantly, the goal of the present study is to motivate a line of inquiry: shifting from manually designed algorithms toward learned algorithms for model editing.
\section{Related Works}
\label{subsec:related_work}
\noindent\textbf{Learned Optimizers.} Neural networks are typically trained using hand-designed optimization algorithms such as gradient descent and its variants, including Adam \citep{kingma2014adam}, AdaGrad \citep{duchi2011adaptive}, and the recently proposed Muon optimizer \citep{jordan2024muon}. A growing body of work investigates whether optimization algorithms can be learned. \citet{3157382.3157543} introduced LSTM-based learned optimizers \citep{hochreiter1997long} and showed that they can outperform hand-designed methods on the tasks for which they are trained. More recently, \citet{chen2023symbolic} formulated algorithm discovery as program search and applied it to discover the Lion optimization algorithm. \citet{li2016learning} reframed learning an optimization algorithm as a reinforcement learning problem and showed that the learned optimizer converges faster than hand-engineered optimizers. Collectively, these works suggest that parameter update rules need not be manually specified and can instead be learned from data. Our work is inspired by the same intuition, but asks a different question: rather than learning how to optimize against a loss function, can an agent learn how to modify a model's behavior using only reward feedback? \\

\noindent\textbf{Model Editing.} A complementary line of work studies post-hoc modification of trained models, often to fix erroneous or undesired behavior. Early work by \citet{sotoudeh2019correcting} formalised neural editing as a satisfiability modulo theory problem. \citet{sinitsin2020editable} introduced the notion of editable training, enabling models to efficiently patch mistakes on individual examples with minimal side effects on unrelated predictions. More recent methods focus on large language models \citep{gangadhar2024model, meng2022mass, meng2022locating,mitchell2022memory}. \citet{gu2024model} showed that model editing methods may result in degradation of general capabilities of the model. Building on these works, we explore a different perspective: instead of manually designing editing algorithms, we investigate whether we can learn to edit models through reinforcement learning. This formulation treats model editing itself as a sequential decision-making problem and opens the possibility of discovering editing strategies from experience.
\section*{Limitations}
While our work suggests a promising research direction, the current framework inherits several limitations commonly associated with reinforcement learning, particularly with respect to scalability. Although our LoRA-inspired mechanism enables modification of high-dimensional layers, fully modifying and optimizing all layers of the model remains infeasible at the current stage, as the enormous action space makes policy optimization highly unstable. Future work should address these limitations and extend the framework through the integration of multi-agent reinforcement learning and game theory. 

\bibliography{latex/0_main}

@article{stable-baselines3,
  author  = {Antonin Raffin and Ashley Hill and Adam Gleave and Anssi Kanervisto and Maximilian Ernestus and Noah Dormann},
  title   = {Stable-Baselines3: Reliable Reinforcement Learning Implementations},
  journal = {Journal of Machine Learning Research},
  year    = {2021},
  volume  = {22},
  number  = {268},
  pages   = {1--8},
  url     = {http://jmlr.org/papers/v22/20-1364.html}
}

@article{schulman2017proximal,
  title={Proximal policy optimization algorithms},
  author={Schulman, John and Wolski, Filip and Dhariwal, Prafulla and Radford, Alec and Klimov, Oleg},
  journal={arXiv preprint arXiv:1707.06347},
  year={2017}
}

@article{lecun1998gradient,
  title={Gradient-based learning applied to document recognition},
  author={LeCun, Yann and Bottou, L{\'e}on and Bengio, Yoshua and Haffner, Patrick},
  journal={Proceedings of the IEEE},
  volume={86},
  number={11},
  pages={2278--2324},
  year={1998},
  publisher={Ieee}
}

@article{lecun1998mnist,
  title={The MNIST database of handwritten digits},
  author={LeCun, Yann},
  journal={http://yann. lecun. com/exdb/mnist/},
  year={1998}
}

@inproceedings{
hu2022lora,
title={Lo{RA}: Low-Rank Adaptation of Large Language Models},
author={Edward J Hu and yelong shen and Phillip Wallis and Zeyuan Allen-Zhu and Yuanzhi Li and Shean Wang and Lu Wang and Weizhu Chen},
booktitle={International Conference on Learning Representations},
year={2022},
url={https://openreview.net/forum?id=nZeVKeeFYf9}
}

@INPROCEEDINGS{9519428,
  author={Bourtoule, Lucas and Chandrasekaran, Varun and Choquette-Choo, Christopher A. and Jia, Hengrui and Travers, Adelin and Zhang, Baiwu and Lie, David and Papernot, Nicolas},
  booktitle={2021 IEEE Symposium on Security and Privacy (SP)}, 
  title={Machine Unlearning}, 
  year={2021},
  volume={},
  number={},
  pages={141-159},
  doi={10.1109/SP40001.2021.00019}}

@INPROCEEDINGS{7163042,
  author={Cao, Yinzhi and Yang, Junfeng},
  booktitle={2015 IEEE Symposium on Security and Privacy}, 
  title={Towards Making Systems Forget with Machine Unlearning}, 
  year={2015},
  volume={},
  number={},
  pages={463-480},
  doi={10.1109/SP.2015.35}}

@misc{jigsaw-toxic-comment-classification-challenge,
    author = {cjadams and Jeffrey Sorensen and Julia Elliott and Lucas Dixon and Mark McDonald and nithum and Will Cukierski},
    title = {Toxic Comment Classification Challenge},
    year = {2017},
    howpublished = {\url{https://kaggle.com/competitions/jigsaw-toxic-comment-classification-challenge}},
    note = {Kaggle}
}

@inproceedings{dixon2018measuring,
author = {Dixon, Lucas and Li, John and Sorensen, Jeffrey and Thain, Nithum and Vasserman, Lucy},
title = {Measuring and Mitigating Unintended Bias in Text Classification},
year = {2018},
isbn = {9781450360128},
publisher = {Association for Computing Machinery},
address = {New York, NY, USA},
url = {https://doi.org/10.1145/3278721.3278729},
doi = {10.1145/3278721.3278729},
booktitle = {Proceedings of the 2018 AAAI/ACM Conference on AI, Ethics, and Society},
pages = {67–73},
numpages = {7},
location = {New Orleans, LA, USA},
series = {AIES '18}
}

@misc{kingma2014adam,
      title={Adam: A Method for Stochastic Optimization}, 
      author={Diederik P. Kingma and Jimmy Ba},
      year={2017},
      eprint={1412.6980},
      archivePrefix={arXiv},
      primaryClass={cs.LG},
      url={https://arxiv.org/abs/1412.6980}, 
}

@misc{jordan2024muon,
  author       = {Keller Jordan and Yuchen Jin and Vlado Boza and Jiacheng You and
                  Franz Cesista and Laker Newhouse and Jeremy Bernstein},
  title        = {Muon: An optimizer for hidden layers in neural networks},
  year         = {2024},
  url          = {https://kellerjordan.github.io/posts/muon/}
}

@inproceedings{3157382.3157543,
author = {Andrychowicz, Marcin and Denil, Misha and Colmenarejo, Sergio G{\'o}mez and Hoffman, Matthew W. and Pfau, David and Schaul, Tom and Shillingford, Brendan and de Freitas, Nando},
title = {Learning to learn by gradient descent by gradient descent},
year = {2016},
isbn = {9781510838819},
publisher = {Curran Associates Inc.},
address = {Red Hook, NY, USA},
booktitle = {Proceedings of the 30th International Conference on Neural Information Processing Systems},
pages = {3988–3996},
numpages = {9},
location = {Barcelona, Spain},
series = {NIPS'16}
}

@article{duchi2011adaptive,
  author  = {John Duchi and Elad Hazan and Yoram Singer},
  title   = {Adaptive Subgradient Methods for Online Learning and Stochastic Optimization},
  journal = {Journal of Machine Learning Research},
  year    = {2011},
  volume  = {12},
  number  = {61},
  pages   = {2121--2159},
  url     = {http://jmlr.org/papers/v12/duchi11a.html}
}

@article{hochreiter1997long,
  title={Long short-term memory},
  author={Hochreiter, Sepp and Schmidhuber, J{\"u}rgen},
  journal={Neural computation},
  volume={9},
  number={8},
  pages={1735--1780},
  year={1997},
  publisher={MIT press}
}

@inproceedings{chen2023symbolic,
author = {Chen, Xiangning and Liang, Chen and Huang, Da and Real, Esteban and Wang, Kaiyuan and Pham, Hieu and Dong, Xuanyi and Luong, Thang and Hsieh, Cho-Jui and Lu, Yifeng and Le, Quoc V.},
title = {Symbolic discovery of optimization algorithms},
year = {2023},
publisher = {Curran Associates Inc.},
address = {Red Hook, NY, USA},
booktitle = {Proceedings of the 37th International Conference on Neural Information Processing Systems},
articleno = {2140},
numpages = {29},
location = {New Orleans, LA, USA},
series = {NIPS '23}
}

@article{li2016learning,
  title={Learning to optimize},
  author={Li, Ke and Malik, Jitendra},
  journal={arXiv preprint arXiv:1606.01885},
  year={2016}
}

@misc{ha2016hypernetworks,
      title={HyperNetworks}, 
      author={David Ha and Andrew Dai and Quoc V. Le},
      year={2016},
      eprint={1609.09106},
      archivePrefix={arXiv},
      primaryClass={cs.LG},
      url={https://arxiv.org/abs/1609.09106}, 
}

@article{jumper2021highly,
  title={Highly accurate protein structure prediction with AlphaFold},
  author={Jumper, John and Evans, Richard and Pritzel, Alexander and Green, Tim and Figurnov, Michael and Ronneberger, Olaf and Tunyasuvunakool, Kathryn and Bates, Russ and {\v{Z}}{\'\i}dek, Augustin and Potapenko, Anna and others},
  journal={nature},
  volume={596},
  number={7873},
  pages={583--589},
  year={2021},
  publisher={Nature Publishing Group UK, London}
}

@article{silver2017mastering,
  title={Mastering chess and shogi by self-play with a general reinforcement learning algorithm},
  author={Silver, David and Hubert, Thomas and Schrittwieser, Julian and Antonoglou, Ioannis and Lai, Matthew and Guez, Arthur and Lanctot, Marc and Sifre, Laurent and Kumaran, Dharshan and Graepel, Thore and others},
  journal={arXiv preprint arXiv:1712.01815},
  year={2017}
}

@book{sutton1998reinforcement,
  title={Reinforcement learning: An introduction},
  author={Sutton, Richard S and Barto, Andrew G and Barto, Andrew},
  volume={1},
  number={1},
  year={1998},
  publisher={MIT press, Cambridge}
}

@article{ouyang2022training,
  title={Training language models to follow instructions with human feedback},
  author={Ouyang, Long and Wu, Jeffrey and Jiang, Xu and Almeida, Diogo and Wainwright, Carroll and Mishkin, Pamela and Zhang, Chong and Agarwal, Sandhini and Slama, Katarina and Ray, Alex and others},
  journal={Advances in neural information processing systems},
  volume={35},
  pages={27730--27744},
  year={2022}
}

@article{meng2022locating,
  title={Locating and editing factual associations in gpt},
  author={Meng, Kevin and Bau, David and Andonian, Alex and Belinkov, Yonatan},
  journal={Advances in neural information processing systems},
  volume={35},
  pages={17359--17372},
  year={2022}
}

@article{meng2022mass,
  title={Mass-editing memory in a transformer},
  author={Meng, Kevin and Sharma, Arnab Sen and Andonian, Alex and Belinkov, Yonatan and Bau, David},
  journal={arXiv preprint arXiv:2210.07229},
  year={2022}
}

@inproceedings{sotoudeh2019correcting,
  title={Correcting deep neural networks with small, generalizing patches},
  author={Sotoudeh, Matthew and Thakur, A},
  booktitle={Workshop on safety and robustness in decision making},
  year={2019}
}

@article{sinitsin2020editable,
  title={Editable neural networks},
  author={Sinitsin, Anton and Plokhotnyuk, Vsevolod and Pyrkin, Dmitriy and Popov, Sergei and Babenko, Artem},
  journal={arXiv preprint arXiv:2004.00345},
  year={2020}
}

@inproceedings{gangadhar2024model,
  title={Model editing by standard fine-tuning},
  author={Gangadhar, Govind Krishnan and Stratos, Karl},
  booktitle={Findings of the Association for Computational Linguistics: ACL 2024},
  pages={5907--5913},
  year={2024}
}

@inproceedings{gu2024model,
  title={Model editing harms general abilities of large language models: Regularization to the rescue},
  author={Gu, Jia-Chen and Xu, Hao-Xiang and Ma, Jun-Yu and Lu, Pan and Ling, Zhenhua and Chang, Kai-Wei and Peng, Nanyun},
  booktitle={Proceedings of the 2024 Conference on Empirical Methods in Natural Language Processing},
  pages={16801--16819},
  year={2024}
}

@inproceedings{mitchell2022memory,
  title={Memory-based model editing at scale},
  author={Mitchell, Eric and Lin, Charles and Bosselut, Antoine and Manning, Christopher D and Finn, Chelsea},
  booktitle={International conference on machine learning},
  pages={15817--15831},
  year={2022},
  organization={PMLR}
}

@article{hort2024bias,
  title={Bias mitigation for machine learning classifiers: A comprehensive survey},
  author={Hort, Max and Chen, Zhenpeng and Zhang, Jie M and Harman, Mark and Sarro, Federica},
  journal={ACM Journal on Responsible Computing},
  volume={1},
  number={2},
  pages={1--52},
  year={2024},
  publisher={ACM, New York, NY}
}

@inproceedings{tian2025identifying,
  title={Identifying and mitigating position bias of multi-image vision-language models},
  author={Tian, Xinyu and Zou, Shu and Yang, Zhaoyuan and Zhang, Jing},
  booktitle={2025 IEEE/CVF Conference on Computer Vision and Pattern Recognition (CVPR)},
  pages={10599--10609},
  year={2025},
  organization={IEEE}
}

@inproceedings{gur2025precise,
  title={Precise in-parameter concept erasure in large language models},
  author={Gur-Arieh, Yoav and Suslik, Clara Haya and Hong, Yihuai and Barez, Fazl and Geva, Mor},
  booktitle={Proceedings of the 2025 Conference on Empirical Methods in Natural Language Processing},
  pages={18997--19017},
  year={2025}
}
\clearpage
\appendix
\section{Appendix}
\subsection{Dataset Splits \& Model Architecture}
\label{subsec:dataset}
\noindent\textbf{Machine Unlearning.} 
We split the MNIST training set \citep{lecun1998gradient, lecun1998mnist} into three subsets: a training set for neural network training (70\%), a training set for policy optimization (15\%), and a validation set for ablation studies and hyperparameter tuning (15\%). We use the standard MNIST test set for all reported evaluations. The model is a CNN consisting of two convolutional layers with 3×3 kernels (16 and 32 filters, respectively), each followed by ReLU activation and 2×2 max-pooling. The extracted feature maps are flattened and passed through three fully connected layers of sizes 128, 32, and 10. In this work, the agent learns to edit the weights of the second fully connected layer, whose parameter matrix has dimensions 128×32. The rank of our LoRA-inspired mechanism was set to 4. The action space was bound between -0.05 and 0.05 for ShiftWorld and 0.0 and 1.0 for MaskWorld for $A$ and $B$.

\noindent\textbf{Bias Mitigation.} We create a balanced subset of 20,000 samples from the Jigsaw Toxic Comment Classification Challenge dataset \citep{jigsaw-toxic-comment-classification-challenge}, which we divide into training, validation, and test sets in a 70/15/15 ratio. We also sample 15,000 examples from the synthetic test set introduced by \cite{dixon2018measuring} and similarly divide them into training, validation, and test sets using a 70/15/15 split. The model consists of an embedding layer followed by fully connected layers, designed to perform text classification. Input text is first passed through an embedding layer with an embedding dimension of 64. The resulting encoded representation is then passed through three fully connected layers with 64, 16, and 2 neurons, respectively. ReLU activation is applied after each hidden layer to introduce non-linearity. In this work, the policy learns to edit the weights of the final fully connected layer, whose parameter matrix has dimensions 16×2. The action space was bound between -0.5 and 0.5 for ShiftWorld and 0.0 and 1.0 for MaskWorld.
\subsection{Hyperparameter Values}
\label{subsec:hyperparameter Values}
All experiments were conducted on CPU. We use the Stable Baselines3 PPO implementation~\citep{stable-baselines3} with the following hyperparameters across all experiments: learning rate $3 \times 10^{-3}$ for MaskWorld and $3 \times 10^{-4}$ for ShiftWorld, $n_{\text{steps}} = 256$, discount factor $\gamma = 0.99$, clipping range $\epsilon = 0.2$, and maximum gradient norm $0.5$. All other parameters are set to library defaults. The total number of timesteps was selected via grid search.

The environment-specific factors for machine unlearning experiments are as follows: episode length $= 1$, number of samples per timestep $= 1024$, $\alpha = 3.0$, and $\beta = -2.0$. The environment-specific factors for bias mitigation experiments are as follows: episode length $= 1$, number of samples per timestep $= 256$, $\alpha = 1.0$, and $\beta = 3.0$. Please refer to the implementation given at \url{https://github.com/shaivimalik/hyperl} for more details.
\subsection{Ablation and Sensitivity Studies}
\label{subsec:ablation}
We use the validation sets described in Appendix~\ref{subsec:dataset} to analyze the impact of several factors: episode length (i.e., the number of actions taken by the agent per episode), number of samples used to calculate the reward at each timestep, the reward coefficients \( \alpha \) and \( \beta \). We perform one-factor-at-a-time ablations, varying one factor while keeping all others fixed at their default values. Default values for the machine unlearning task:
\texttt{episode\_length}=1, \texttt{batch\_size}=1024, \( \alpha=3.0 \), \( \beta=2.0 \). Tables ~\ref{table:appen_epi}, ~\ref{table:alpha}, ~\ref{table:beta} and ~\ref{table:batch} summarize the corresponding ablation results. 
The total number of training timesteps was held constant across all studies. Consequently, increasing the episode length effectively reduces the number of episodes seen during training, which may give the appearance of degraded policy performance. Given a sufficient number of training timesteps, we expect the learned policies to converge regardless of episode length.
\begin{table*}
\centering
\small
\begin{tabularx}{\textwidth}{l X X X X X}
\hline
\textbf{Environment} &
\textbf{Episode Length} &
\textbf{Original Model Accuracy on Retain Set} &
\textbf{Original Model Accuracy on Forget Set} &
\textbf{Modified Model Accuracy on Retain Set} &
\textbf{Modified Model Accuracy on Forget Set} \\
\hline

MaskWorld & 1 & 0.912 $\pm$ 0.016 & 0.914 $\pm$ 0.046
& 0.932 $\pm$ 0.005 & 0.000 $\pm$ 0.000\\
MaskWorld & 2 & 0.912 $\pm$ 0.016 & 0.914 $\pm$ 0.046
& 0.919 $\pm$ 0.007 & 0.000 $\pm$ 0.000\\
MaskWorld & 4 & 0.912 $\pm$ 0.016 & 0.914 $\pm$ 0.046
& 0.861 $\pm$ 0.047 & 0.000 $\pm$ 0.000\\

ShiftWorld & 1 & 0.912 $\pm$ 0.016 & 0.914 $\pm$ 0.046
& 0.932 $\pm$ 0.005 & 0.000 $\pm$ 0.000\\
ShiftWorld & 2 & 0.912 $\pm$ 0.016 & 0.914 $\pm$ 0.046
& 0.888 $\pm$ 0.030 & 0.000 $\pm$ 0.000\\
ShiftWorld & 4 & 0.912 $\pm$ 0.016 & 0.914 $\pm$ 0.046
& 0.598 $\pm$ 0.141 & 0.000 $\pm$ 0.000\\

\hline
\end{tabularx}

\caption{
Effect of Episode Length on Performance of Learned Policies Across Environments.
}
\label{table:appen_epi}
\end{table*}

\begin{table*}
\centering
\small
\begin{tabularx}{\textwidth}{l l X X X X}
\hline
\textbf{Environment} &
\textbf{$\alpha$} &
\textbf{Original Model Accuracy on Retain Set} &
\textbf{Original Model Accuracy on Forget Set} &
\textbf{Modified Model Accuracy on Retain Set} &
\textbf{Modified Model Accuracy on Forget Set} \\
\hline

MaskWorld & 1 & 0.912 $\pm$ 0.016 & 0.914 $\pm$ 0.046
& 0.931 $\pm$ 0.006 & 0.000 $\pm$ 0.000\\
MaskWorld & 3 & 0.912 $\pm$ 0.016 & 0.914 $\pm$ 0.046
& 0.932 $\pm$ 0.005 & 0.000 $\pm$ 0.000\\
MaskWorld & 10 & 0.912 $\pm$ 0.016 & 0.914 $\pm$ 0.046
& 0.933 $\pm$ 0.005 & 0.000 $\pm$ 0.000\\

ShiftWorld & 1 & 0.912 $\pm$ 0.016 & 0.914 $\pm$ 0.046 & 0.930 $\pm$ 0.004 & 0.000 $\pm$ 0.000\\
ShiftWorld & 3 & 0.912 $\pm$ 0.016 & 0.914 $\pm$ 0.046 & 0.932 $\pm$ 0.005 & 0.000 $\pm$ 0.000\\
ShiftWorld & 10 & 0.912 $\pm$ 0.016 & 0.914 $\pm$ 0.046 & 0.931 $\pm$ 0.005 & 0.000 $\pm$ 0.000\\

\hline
\end{tabularx}

\caption{
Effect of $\alpha$ on Performance of Learned Policies Across Environments.
}
\label{table:alpha}
\end{table*}

\begin{table*}
\centering
\small
\begin{tabularx}{\textwidth}{l l X X X X}
\hline
\textbf{Environment} &
\textbf{$\beta$} &
\textbf{Original Model Accuracy on Retain Set} &
\textbf{Original Model Accuracy on Forget Set} &
\textbf{Modified Model Accuracy on Retain Set} &
\textbf{Modified Model Accuracy on Forget Set} \\
\hline

MaskWorld & -1 & 0.912 $\pm$ 0.016 & 0.914 $\pm$ 0.046
& 0.931 $\pm$ 0.006 & 0.000 $\pm$ 0.000\\
MaskWorld & -2 & 0.912 $\pm$ 0.016 & 0.914 $\pm$ 0.046
& 0.932 $\pm$ 0.005 & 0.000 $\pm$ 0.000\\
MaskWorld & -5 & 0.912 $\pm$ 0.016 & 0.914 $\pm$ 0.046
& 0.931 $\pm$ 0.006 & 0.000 $\pm$ 0.000\\

ShiftWorld & -1 & 0.912 $\pm$ 0.016 & 0.914 $\pm$ 0.046 
& 0.932 $\pm$ 0.004 & 0.000 $\pm$ 0.000\\
ShiftWorld & -2 & 0.912 $\pm$ 0.016 & 0.914 $\pm$ 0.046 
& 0.932 $\pm$ 0.005 & 0.000 $\pm$ 0.000\\
ShiftWorld & -5 & 0.912 $\pm$ 0.016 & 0.914 $\pm$ 0.046 
& 0.928 $\pm$ 0.003 & 0.000 $\pm$ 0.000\\

\hline
\end{tabularx}

\caption{
Effect of $\beta$ on Performance of Learned Policies Across Environments.
}
\label{table:beta}
\end{table*}

\begin{table*}
\centering
\small
\begin{tabularx}{\textwidth}{l l X X X X}
\hline
\textbf{Environment} &
\textbf{Number of Samples} &
\textbf{Original Model Accuracy on Retain Set} &
\textbf{Original Model Accuracy on Forget Set} &
\textbf{Modified Model Accuracy on Retain Set} &
\textbf{Modified Model Accuracy on Forget Set} \\
\hline

MaskWorld & 256 & 0.969 $\pm$ 0.003 & 0.976 $\pm$ 0.008 & 0.972 $\pm$ 0.004 & 0.000 $\pm$ 0.000\\
MaskWorld & 512 & 0.947 $\pm$ 0.007 & 0.956 $\pm$ 0.026 & 0.961 $\pm$ 0.002 & 0.000 $\pm$ 0.000\\
MaskWorld & 1024 & 0.912 $\pm$ 0.016 & 0.914 $\pm$ 0.046 & 0.932 $\pm$ 0.005 & 0.000 $\pm$ 0.000\\
ShiftWorld & 256 & 0.968 $\pm$ 0.004 & 0.976 $\pm$ 0.008 & 0.944 $\pm$ 0.015 & 0.008 $\pm$ 0.008\\
ShiftWorld & 512 & 0.947 $\pm$ 0.007 & 0.956 $\pm$ 0.026 & 0.958 $\pm$ 0.005 & 0.000 $\pm$ 0.000\\
ShiftWorld & 1024 & 0.912 $\pm$ 0.016 & 0.914 $\pm$ 0.046 & 0.932 $\pm$ 0.005 & 0.000 $\pm$ 0.000\\

\hline
\end{tabularx}

\caption{
Effect of Number of Samples Used to Calculate the Reward at Each Timestep on Performance of Learned Policies Across Environments.
}
\label{table:batch}
\end{table*}

\subsection{Statistical Significance Analysis}
To assess the statistical significance of our results, we performed the paired \textit{t}-test and the Wilcoxon signed-rank test to compare the performance of the models. Tables ~\ref{tab:bias_stats} and ~\ref{tab:unlearn_stats} present the corresponding \textit{p}-values. The results show that the observed differences are statistically significant at the 0.05 significance level (\textit{p} < 0.05). 

\begin{table*}[t]
\centering
\small
\resizebox{\textwidth}{!}{%
\begin{tabular}{lcccc}
\toprule
\textbf{Comparison} & \textbf{Mean Difference} & \textbf{95\% CI} & \textbf{Paired \textit{t}-test (\textit{p})} & \textbf{Wilcoxon signed-rank test (\textit{p})} \\
\midrule
MaskWorld vs.\ Original (Unbiased Dataset)  & $+7.24$ & [0.0560, 0.0888]    & 0.0000 & 0.0001 \\
MaskWorld vs.\ Original (General Dataset)   & $-4.95$ & [$-$0.0686, $-$0.0303] & 0.0001 & 0.0001 \\
ShiftWorld vs.\ Original (Unbiased Dataset) & $+7.73$ & [0.0608, 0.0938]    & 0.0000 & 0.0001 \\
ShiftWorld vs.\ Original (General Dataset)  & $-4.77$ & [$-$0.0590, $-$0.0363] & 0.0000 & 0.0001 \\
\bottomrule
\end{tabular}%
}
\caption{Results of paired \textit{t}-tests and Wilcoxon signed-rank tests comparing performance of the RL-edited models (MaskWorld and ShiftWorld) with the original model on the bias mitigation task. Mean performance differences are reported in percentage points.}
\label{tab:bias_stats}
\end{table*}

\begin{table*}[t]
\centering
\small
\resizebox{\textwidth}{!}{%
\begin{tabular}{lcccc}
\toprule
\textbf{Comparison} & \textbf{Mean Difference} & \textbf{95\% CI} & \textbf{Paired \textit{t}-test (\textit{p})} & \textbf{Wilcoxon signed-rank test (\textit{p})} \\
\midrule
MaskWorld vs.\ Original (Forget Dataset)  & $-90.73$ & [$-$0.9313, $-$0.8834] & 0.0000 & 0.0001 \\
MaskWorld vs.\ Original (Retain Dataset)  & $+1.34$  & [0.0048, 0.0219]    & 0.0047 & 0.0004 \\
ShiftWorld vs.\ Original (Forget Dataset) & $-90.72$ & [$-$0.9312, $-$0.8833] & 0.0000 & 0.0001 \\
ShiftWorld vs.\ Original (Retain Dataset) & $+1.40$  & [0.0057, 0.0224]    & 0.0030 & 0.0007 \\
\bottomrule
\end{tabular}%
}
\caption{Results of paired \textit{t}-tests and Wilcoxon signed-rank tests comparing performance of the RL-edited models (MaskWorld and ShiftWorld) with the original model on the machine unlearning task. Mean performance differences are reported in percentage points.}
\label{tab:unlearn_stats}
\end{table*}
\end{document}